\documentclass{nime-alternate} 

\usepackage{anonymize} 		   

\usepackage[utf8]{inputenc}

\begin{document}

\conferenceinfo{NIME'20,}{July 21-25, 2020, Royal Birmingham Conservatoire, ~~~~~~~~~~~~ Birmingham City University, Birmingham, United Kingdom.}
\title{Mechatronics-Driven Musical Expressivity for Robotic Percussionists}

%
%
%
\label{key}
%

\numberofauthors{1} 
%
\author{
%
%
\alignauthor
\anonymize{Ning Yang, Richard Savery, Raghavasimhan Sankaranarayanan,\\ Lisa Zahray, Gil Weinberg}\\
       \affaddr{\anonymize{Georgia Tech Center for Music Technology}}\\
       \affaddr{\anonymize{Atlanta, USA}}\\
       \email{\anonymize{nyang8@gatech.edu}}
}



\maketitle
\begin{abstract}
Musical expressivity is an important aspect of musical performance for humans as well as robotic musicians. We present a novel mechatronics-driven implementation of Bru-shless Direct Current (BLDC) motors in a robotic marimba player, named \anonymize{Shimon}, designed to improve speed, dynamic range (loudness), and ultimately perceived musical expressivity in comparison to state-of-the-art robotic percussionist actuators. In an objective test of dynamic range, we find that our implementation provides wider and more consistent dynamic range response in comparison with solenoid-based robotic percussionists. Our implementation also outperforms both solenoid and human marimba players in striking speed. In a subjective listening test measuring musical expressivity, our system performs significantly better than a solenoid-based system and is statistically indistinguishable from human performers.
\end{abstract}




\section{Introduction}
The ability to perform music expressively is often considered outside the capability of robot and computer music systems \cite{dobrian2006nime}.
Multiple studies show expressivity can be achieved through subtle manipulation of volume and micro timings \cite{baraldi2006communicating, canazza1996sonological,mion2008score}.
This paper proposes a new implementation of Brushless Direct Current (BLDC) motors in a robotic musician platform named \anonymize{Shimon}, focusing on capturing human musical expressivity through accurate micro-timing control and high-resolution consistent dynamic range. Unlike previous actuators used in robotic percussionists such as solenoids and linear motors, our implementation promises to provide accurate and expressive performance that is indistinguishable from  expressive human performance. The system was evaluated in comparison with four marimba players using extensive listening tests to identify its ability to equal human-level musical expressivity.


%

\pagebreak
    \begin{figure}[h]
        \centering
        \includegraphics[width =8cm]{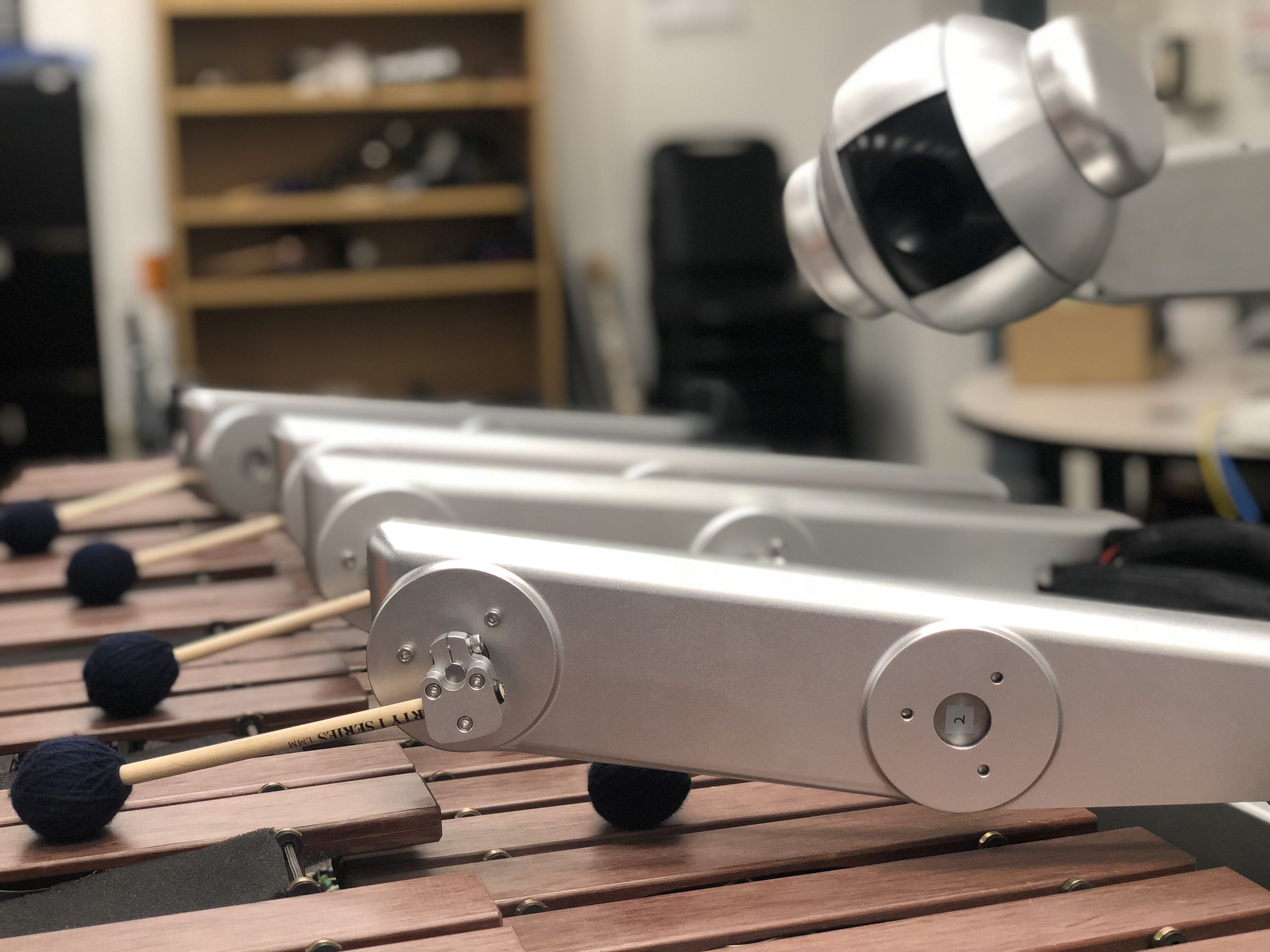}
        \caption{\anonymize{Shimon's} new design - eight new BLDC motors have replaced the robot's previous solenoid driven design}
        \label{fig:shimonarm}
    \end{figure}
\section{Related Work}    
Our work is situated in an interdisciplinary space that includes research in Robotic Musicianship (RM) and human performance expressivity. Research in RM addresses the construction of machines that can produce sound, analyze and generate musical input, and interact with humans in a musically meaningful manner \cite{bretan2016survey}. There are two primary research areas that constitute the field of RM: 1). Machine Musicianship - the development of algorithms and cognitive models of music perception, composition, performance, and theory \cite{rowe2004machine} and 2). Musical Mechatronics - the study and construction of physical devices that generate sound through mechanical means \cite{kapur2005history}.  While recent works in RM have led to significant advancements in Machine Musicianship including aspects such as machine improvisation \cite{bretan2016unit}, group synchronization and anticipation \cite{hoffman2010synchronization, cicconet2013human}, and turn taking \cite{weinberg2006jam}, little progress has been made to improve the acoustic expressivity of robotic musicians through advancements in Musical Mechatronics. A number of notable Musical Mechatronics efforts have addressed wind instruments  \cite{dannenberg2005mcblare, solis2010development}, and string instruments \cite{singer2004lemur, kusuda2008toyota}. Several robotic systems have been designed to capture human expressivity, focusing on wrist and arm movements. The {\it Cog} robot from MIT, for example, uses oscillators for smoother rhythmic arm control \cite{williamson1999robot,williamson1999designing}. Another approach to smooth motion control is to use hydraulics as done by Mitsuo Kawato in his humanoid drummer \cite{atkeson2000using}.  However, most of the efforts to develop robotic percussionists have used solenoid actuation \cite{maes2011man, kapur2011karmetik,  singer2004lemur}, which cannot support human-level dynamic range and millisecond micro-timing  -  key required elements for musical expressivity. Baraldi et. al have shown that both musicians and non-musicians identify expressivity with only a few low level parameters. From a performance perspective these are micro-timing and volume control, while others are largely dictated by the composition itself \cite{baraldi2006communicating}. Canazza et. al \cite{canazza1996sonological} and Mion \cite{mion2008score} demonstrated the importance of notes timing and variations in amplitude for perceived expression.

In our own previous work, we attempted to advance mech-atronics-driven expressivity for robotic percussionists by using actuators such as a linear motor that drove one of the arms of the robotic percussionist Haile \cite{weinberg2005haile} and a gimbal motor that was used for the Robotic Drumming Prosthetic (RDP) \cite{bretan2016robotic}.  Haile's linear motor provided wider and more consistent dynamic range than possible with solenoids \cite{kapur2007comparison}. However, this came at the expense of striking speed, which was limited to 6 Hz and could not support more subtle striker micro-timings. The gimbal motor used for the RDP, on the other hand, was able to reach much higher speed (20 Hz) and control a variety of bouncing profiles using PID control \cite{gopinath2016generative}. However, due to its design, the RDP could not control wide and consistent dynamic range and therefore could not support rich musical expressivity. In an effort to significantly improve both dynamic range and micro-timing control for robotic percussionists, we redesigned our solenoid-based marimba-playing robotic musician \anonymize{Shimon} \anonymize{\cite{hoffman2010shimon}}. For the work described in this paper, we replaced \anonymize{Shimon's} solenoids with BLDC motors \cite{pillay1989modeling}, which hitherto have not been used for RM, in an effort to achieve human level musical expressivity through wide dynamic range and high resolution and consistent micro-timing control.

\begin{figure}[h]
    \centering
    \includegraphics[width =8cm]{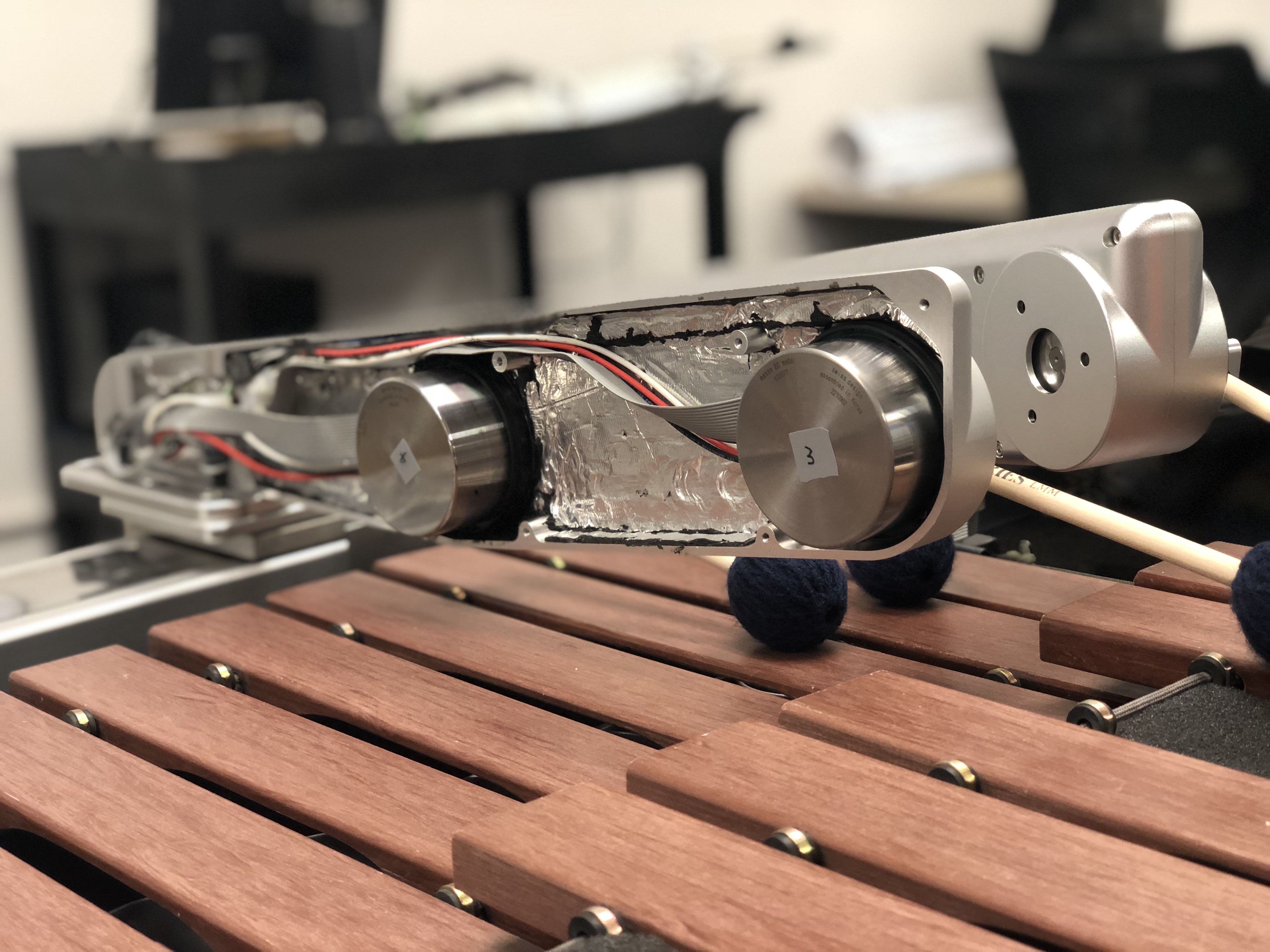}
    \caption{An inside view of one of \anonymize{Shimon's} new arms  - each arm consists of two BLDC motors, two clamps and sound deadening mat}
    \label{fig:shimonmotor}
\end{figure}

\section{Motivation and Approach}

Informed by the related work described above, our motivation for this project was to identify, implement, and test a mechanical actuator that would allow for wide and consistent dynamic range control and high playing speeds that could facilitate human-level performance expressivity. In addition to conducting objective evaluation of our implementation in comparison to state-of-the-art robotic percussionist actuators, we also performed subjective evaluation, comparing the musical outcome of our implementation to human performance. To address this motivation, we identified and implemented BLDC motors in our robotic marimba musician named \anonymize{Shimon}\anonymize{\cite{hoffman2010shimon}}. We hypothesize that the motor's fast response rate, PID-control feedback loop, fast oscillation and precise motion control could enable larger dynamic range and playing speeds than can be achieved by state-of-the-art solutions. Moreover, BLDC motors are known for their high torque-to-weight ratio, low noise levels, and long lifespan \cite{bldc2003}, which makes them a suitable choice for our system.

\section {Design and Implementation}

\subsection{Platform}

\begin{figure}
    \centering
    \includegraphics[width =4cm]{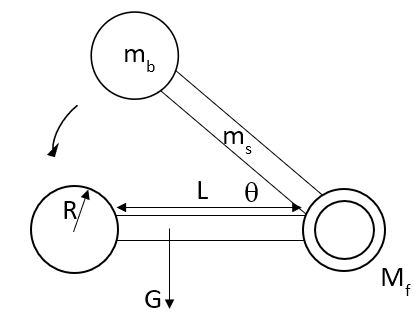}
    \caption{\anonymize{Shimon's} new striker consists of a mallet and a motor shaft. The mallet is modeled as a ball with radius $R$ and a rod with length $L$}
    \label{fig:striker_model}
\end{figure}


The first design requirement we addressed was to replace \anonymize{Shimon's} old solenoid-driven strikers with BLDC motors while maintaining the original design of the robot's arms. To fit into \anonymize{Shimon's} arms, the actuator had to be less than 65 millimeters in diameter, and 40 millimeters in depth. At nominal output, the solenoid provides sufficient loudness. We therefore aimed to select a motor that matched the solenoid's maximum applied torque on the marimba, measured to be 0.3125 Nm.  

In order to assess whether the motor specification fit these requirements, we analyzed a simple 1 degree of freedom model shown in Figure \ref{fig:striker_model}. The mallet was modeled as a combination of a ball and a rod. The ball had radius $R$ and mass $m_b$, and the rod had length L and mass $m_s$. The mallet rotates with the pivot point at the center of the motor with angle $\theta$. For downstroke, the net torque on the striker $M_d$ with motor torque output $M_m$ is,

\begin{equation}
M_d =M_m+M_g-M_f
\end{equation}

where $M_f$ can be found by torque constant $k_m$ and no-load current $I_o$, both provided by the motor manufacturer. The required output torque by the motor is

\begin{equation}
\label{eq:torque}
M_m =M_d-[m_b(R+L)+m_s \frac{L}{2}]sin(\theta)g-k_mI_o
\end{equation}
where the largest torque happens right at the instance of hit.

 We used these calculations to evaluate whether a motor can achieve the desired downstroke torque $M_d$ of 0.3125 Nm. Our resulting choice is Maxon's EC-60 flat motor, which can be seen in Figure \ref{fig:shimonarm}. According to the motor's specifications and Equation \ref{eq:torque}, it requires an output torque $M_m$ of 0.316 Nm to meet the downstroke torque requirement. The EC-60 exceeds this requirement with a nominal torque of 0.319 Nm. It additionally meets our size requirements, with a 65-millimeter diameter and 38-millimeter depth. 
 
 The new system maintained the previous arm design, although the particular motor we chose did require interior redesign of the arms. As shown in Figure \ref{fig:shimonmotor}, each of the aluminum arms is comprised of two mallets, which are clamped directly onto the shafts of two Maxon EC-60 flat motors. The arms' inner walls are attached with sound deadening mat from Noico to damp any noise between the motor and the arm case. Similarly to the old system, the lower striker in each arm was placed to play the white keys of the marimba and the top four strikers were placed in a position to play the row of black keys. The commercial IAI linear actuator, which controls longitude direction of the arms, remained unchanged in the new system.

\subsection{Motor Implementation}
The eight Maxon E-60 motors are controlled by Maxon EPOS4 50/8 PID controllers and communicate through Ethernet with Control Automation Technology (EtherCAT) protocol. The motor control is executed through Beckhoff's TwinCat3 programmable logic controller (PLC) system, operated under Beckhoff's C6015 Industrial PC (IPC). The first controller in the daisy chain connects to the Beckhoff IPC that works as an EtherCAT master, as shown in figure \ref{fig:dataflow}. 

\begin{figure}
    \centering
    \includegraphics[width =8.5cm]{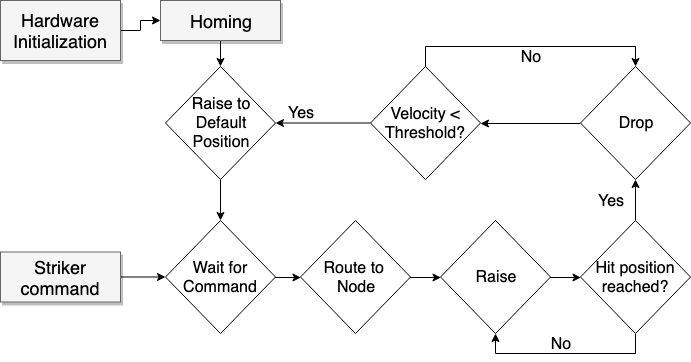}
    \caption{Striker Activation Algorithm}
    \label{fig:algorithm}
\end{figure}

PID values are determined when mallets are clamped to the motor shaft, using an auto tuning function provided by Maxon EPOS studio, a commercially available software to assist the initialization of Maxon motors. Cyclic Synchronous Position Mode (CSP) is used as primary operation mode through Beckhoff TwinCat3 PLC system to achieve tick synchronization in real-time. Both acceleration and position need to be specified in the profile of motion command. All eight motors can then be triggered simultaneously by ticks when \anonymize{Shimon} plays multiple note chords. Since BLDC motors do not hold their positions when the system is not powered, we programmed the PLC system to home motor encoders during initialization to a default position above the marimba (see details  in Figure \ref{fig:algorithm}). 

We have chosen MIDI to communicate between the PLC and the motor controller, since MIDI is a well established protocol for musical instrument communication that is commonly used by musicians to compose for and interact with \anonymize{Shimon}. The angular acceleration defined in the motion command is linearly mapped with MIDI velocity to provide smooth increase in striking volume. The MIDI standard does not specify how to interpret velocity, instead allowing manufactures to map in whatever way is suited to an instruments design \cite{MIDI}. In order to follow musical tempo and to control micro-timing while playing expressively, the travel time for each mallet stroke must remain constant. We therefore linearly interpolated the position values with the accelerations. The value of MIDI velocity can therefore fully define a motion profile. A threshold for MIDI velocity was identified through experimentation to sense whether the mallets hit the marimba surface. Once the threshold is reached, the PLC system sends out another motion command to lift the mallets back up to the default position and wait for the next note. A velocity monitoring command block is built into the PLC system to compensate for real time errors and to maintain stable acoustic sound quality.


\subsection {Communication and Data Flow}

\begin{figure}
    \centering
    \includegraphics[width=8.5cm]{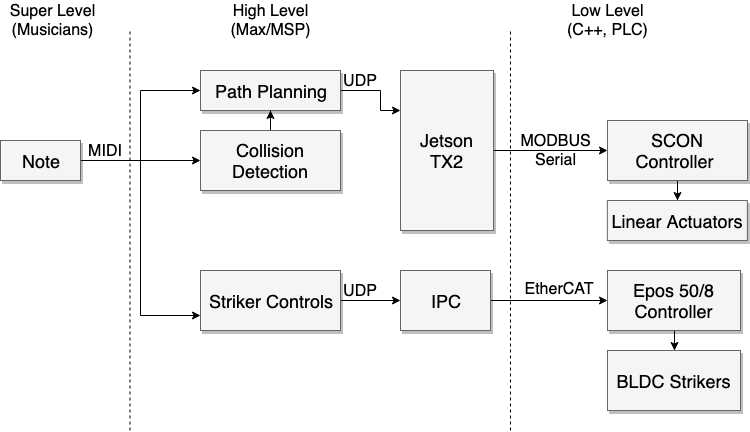}
    \caption{System Communication and Data flow}
    \label{fig:dataflow}
\end{figure}
    
The MIDI instructions are transformed into lower level actuator commands, as the note information is sent synchrono-usly
 to both the linear actuator and the strikers via UDP. UDP messages that are sent from Max/MSP are received by a Nvidia Jetson TX2 and the IPC, as shown in figure \ref{fig:dataflow}. The Nvidia Jetson TX2 was used to allow for future deep learning capabilities, as used in other robotic music platforms \cite{savery2019establishing,savery2019finding}. A motion control library written in C++ executes the motion commands that control the linear actuators. The note's pitch information is linearly translated to sliding movements of the actuators to map the pitches to the respective physical marimba key positions. These key positions are customized to the marimba used and are manually defined. Trajectories are then handled by IAI’s SCON controller. MIDI velocity information is converted to BLDC motor acceleration and position. They are executed by PLC's motion control command, while the EPOS4 motor controllers calculate  the trajectories based on the executed command. This allows for MIDI commands to be specified using musical software such as Logic Pro X and MAX/MSP on a Macbook. Therefore, musicians working with \anonymize{Shimon} do not need to communicate with its low-level control, and focus only on their musical programming.

\section{Experiments}
Three experiments were designed to evaluate the speed, dynamic range, and perceived expression of our BLDC-driven robotic percussionist implementation. Since it has been shown that expressivity can be communicated using  just one note \cite{baraldi2006communicating}, the experiments were designed to use a single motor on a single note. This approach also allowed us to better control for variables and to streamline the experiment protocol.  

\subsection{Research Question and Hypotheses}

 Our goals for the system were to exceed the speed and micro-timing resolution of state-of-art robotic or human percussionists, to support a consistent and wide dynamic range control, and to play expressively in a manner that would be indistinguishable from a human performer. To test whether these goals were achieved, we designed experiments to compare the solenoid, BLDC motor, and human marimba players based on the following hypotheses: 


\begin{enumerate}
    \item Dynamic Range and Speed Capabilities
        \begin{enumerate}
        \item The BLDC motor will be able to perform with a wider and more stable dynamic range than a solenoid when striking on a marimba.
        \item The BLDC motor will outperform both solenoid striker and human players in striking speed on a marimba. 
        \item The BLDC motor striker will be able to reproduce the volume and speed range played by a human or solenoid with equal or greater consistency
        \end{enumerate}
        
    \item Capturing Human expressivity
        \begin{enumerate}
        \item In a listening test, subjects will not be able to recognize if a musical excerpt is played by the our system or a skilled performer
        \end{enumerate}

\end{enumerate}

\section{Methodology}
To evaluate these hypotheses we conducted both objective and subjective experiments. 

\subsection{Objective Experiment Design}
For each objective measure we recorded solenoids, BLDC motors and a human player completing identical tasks.
\subsubsection {Dynamic Range}
Both solenoids and BLDC motors were programmed to play single C2 note at 60 bpm on the marimba, with ascending MIDI velocity of step 1 from 1 to 127 for every 6 strokes. The MIDI notes are triggered by Apple Logic X Digital Audio Workstation.
Each of the marimba players were asked to play single C2 note at 60 bpm on the marimba for 3 different trials. 

\begin{itemize}
\item Play loudest possible for 6 strokes.
       \item Play softest possible for 6 strokes.
       \item Play ascending five dynamic levels from softest to loudest, with 6 strokes per level.
\end{itemize}
An SPL meter records the sound level change throughout all the strokes.

\subsubsection{Speed}
Both solenoids and BLDC motors were programmed to play single C2 note for 3 seconds with system defined speed. Set speed was gradually increased for each trial until either fail of hit or power cutoff. Each of the marimba players were asked to play the note C2 as fast as possible for 3 seconds with three different dynamic requirements. The first trial had no constraint in volume; the second trial required the player to play as soft and as fast as possible; the third trial asked the player to play as loud and as fast as possible.

\subsection {Subjective Experiment Design}
We recorded  four marimba players performing twelve, four measure long expressive excerpts on a single note, collected from Belson's Modern Reading Text \cite{bellson1963modern}. The players were allowed to practice each phrase as long as desired. Each phrase was then recreated for the solenoid and BLDC motors using Abelton Live's audio to midi, followed by human verification.

Musical expressivity was then evaluated using a 30-minute listening test administrated through Qualtrics. We collected results from 21 participants, 14 of which listened in a controlled environment through headphones, while the other 7 listened externally through headphones of their choice. Participants were gathered with high-school level music training or above. 

The process began with a page introduction to musical expressivity, through a collection of quotes, following methodologies from \cite{dobrian2006nime}. This was followed by 36 iterations of the question: Which sample is more musically expressive? We compared solenoid, BLDC and human excerpts in pair-wise configurations so as to have direct comparison between each performance type. Excerpts and questions were randomly ordered. The questionnaire ended with three open-ended questions allowing participants to comment on any trends they noted, any difficulties comparing excerpts and a description of their musical education. 


\subsection{Apparatus}

\subsubsection {Recording Specification}
    
    Audio was recorded with three SM-57 microphones, positioned at the center point beneath each key, 40 centimeters above the ground. The microphones were sent through a Focusrite Scarlett 18i20, with a set calibrated pre-amp volume.  

\subsubsection {Dynamic Range Measurement}

Dynamic ranges were determined through R8080 type 2 sound pressure level (SPL) meter by REED instruments. The SPL meter was calibrated at 94 dB with REED R8090 sound level calibrator and placed 30 centimeters away, pointing towards the sound source. Frequency weighting was set to dBC to capture peak acoustic sound levels and time weighting indicator was set to SLOW to record averaged Decibel rate. Data was recorded on a Windows 10 computer running REED Datalogger Software with sampling rate of 500ms per data sample.
    
\subsubsection {Speed Measurement}

    Mallet movement was recorded through a slow motion video recording with 240 frames per second by an iPhone X. Tracker video analysis and modeling tool was used for motion tracking. Speed was acquired by obtaining the number of periodic movements over fixed time span.


    \begin{figure*}
        \centering
        \includegraphics[width =16cm]{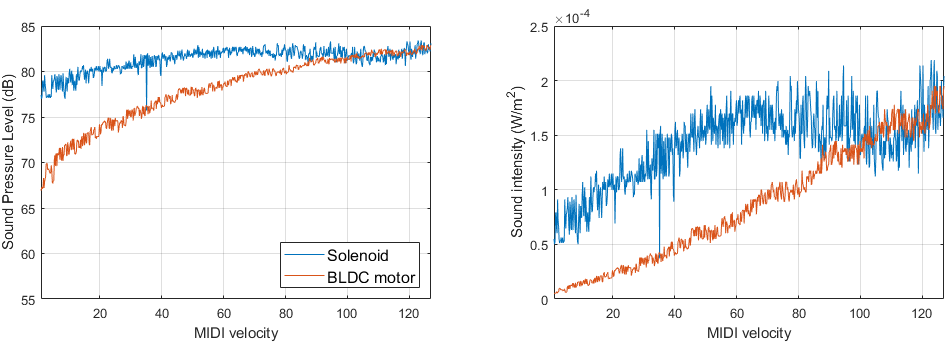}
        \caption{Comparison of SPL and sound intensity of solenoid and BLDC motor in response to MIDI velocity range of 1 to 127}
        \label{fig:comparisonDR}
    \end{figure*}


\section{Results}

\subsection{Objective Experiment Results}

Figure \ref{fig:comparisonDR} presents a comparison between the dynamic range achieved by Shimon's previous solenoid system and our new BLDC motor implementation in response to MIDI velocity from 1 to 127. Room noise was measured as 55 dB. As MIDI velocity increased, logarithmic curves in loudness were observed for both actuators. However, the dynamic range achieved by the solenoid was 73 dB to 83 dB, compared to the larger 57 dB to 83 dB range for the BLDC motor. The solenoid also demonstrated larger fluctuation in sound level in comparison to the BLDC motor. Sound intensity was calculated based on sound pressure level and a reference intensity of 20 micropascals. The result exhibited a linear correlation between sound intensity and MIDI velocity for the BLDC motor through the full dynamic range. The solenoid followed linear correlation at lower dynamic range as well, but failed after MIDI velocity reached 80. The figure additionally shows lower stability in the solenoid's sound intensity control, as indicated by less consistent values. The linear interpolation achieved a norm of residual of 6.899e-05, and showed an increase of 1.4472e-06 $W/m^2$ in sound intensity for each step increase in MIDI velocity.

\subsubsection {Speed}
The solenoid reached up to 8.3 Hz in striking speed using the old striker system, before the actuator failed to hit the marimba surface. Our BLDC motor reached a speed of 32.9 Hz, before the over-current protection cut off the motor controller for safety purposes. In comparison, the average fastest speed recorded from 4 different human marimba players was 7.1 Hz. No significant maximum speed levels in different dynamic ranges were observed among human players.

\subsection{Subjective Experiment Results}
\begin{figure}[h]
        \centering
        \includegraphics[width =8cm]{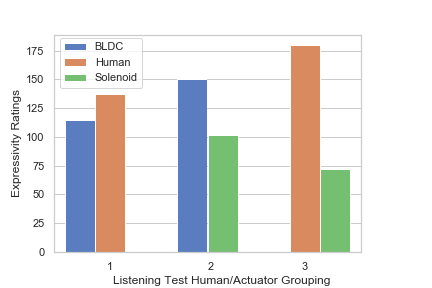}
        \caption{Participant Expressivity Ratings}
        \label{fig:expressive}
    \end{figure}

We conducted a binomial test to compare each grouping of motors and humans (see figure \ref{fig:expressive}) with maximum rating of 252 positive results. The solenoid performed much worse than the human, with ratings of 180 positive to the human, over 72 to the solenoid. This resulted in $p= 1.3e^{-0.9}$. The BLDC motor also performed better than the solenoid, however the value was much closer at $p= 0.0029$. Comparing the human and BLDC motor results, the human excerpts received 137 choices as more expressive over the BLDC motor's 115. However this result was not significant $p=0.19$, failing to reject the null hypothesis. 

\section{Discussion}
Our experiments demonstrated that the BLDC motor provided wider dynamic range than the solenoid. Since the overall size and power are similar between the solenoid and the BLDC motor, it is expected to observe similar maximum sound level for both actuators. The advantage of BLDC motor in respect to dynamic range was more noticeable on the lower range and stability. The BLDC motor provided a full dynamic range consistently and with high resolution from room noise to the system threshold. The solenoid on the other hand had lowest sound level of 74 dB. We associated this performance with the motor's higher response rate and feedback control over both position and acceleration. The captured audio from the solenoids oscillated throughout the whole dynamic range, which could be explained by its lack of feedback control that leads to unstable output torque. On the other hand, our BLDC motor implementation using PID controllers ensured much more stable output torque and position, which led to smooth dynamic range with higher resolution. Both BLDC motor and solenoid outperformed human players in striking speed. However, \anonymize{Shimon's} 8.3 Hz solenoids played relatively slower than other solenoid systems utilized in related work such as Kapur Fingers (14.28 Hz) and Trimpin Hammer (18.18 Hz), although these faster solenoid systems also reported lack of volume controls \cite{kapur2007comparison}. At maximum play speed of 32.9 Hz, the new BLDC motor system supports distinctive micro timing that we associate with enhanced music expressivity.

In our subjective test, most of the 21 participants could not differentiate the excerpts played by humans and \anonymize{Shimon} with BLDC motor, but the excerpts played by humans and solenoid were distinguishable. We explain this by the BLDC's mimetic control in volume and micro timing, which led \anonymize{Shimon} to imitate excerpts from human players indistinguishably. Even though the human excerpts received higher votes in expressivity than those played with the BLDC motor, the result was not significant. Between excerpts of BLDC motor and solenoid, most of the participants voted for the BLDC motor in musical expressivity. Compared with the solenoid system, the new BLDC motor system was able to musically replicate human expressivity in performance with high resolution control in dynamic range, speed and timing.

\section{Future Work}
 
In future work, we plan to further evaluate our implementation of BLDC motors in a more comprehensive musical context, including reproducing full melodic phrases rather than just one-note rhythmic excerpts. Building on our preliminary work on anticipatory visual cues in robotic musicianship, \cite{hoffman2011interactive}, we also plan to take advantage of the accurate position control of the BLDC motor to implement and evaluate novel anticipatory visual cues for human-robot musical interaction. We anticipate that the integration of detailed motion design with both BLDC motors and \anonymize{Shimon's} linear actuator would lead to improved synchronization and coordination between robotic musicians and humans, leading to even more expressive and engaging musical performances.


\section{Acknowledgments}
Our thanks to the marimbists who participated in the experiment; Nichole Beck, Nicholas Farris, Khalil Keyton and Alexis Wilson.

\section{Ethical Standards}
Participants gave informed consent as described in IRB Protocol \anonymize{H19364 issued by Georgia Institute of Technology.}

%
\bibliographystyle{abbrv}

	\bibliography{nime-references}

%

\end{document}